\documentclass{article}

\usepackage[preprint,nonatbib]{neurips_2023}

\usepackage[bookmarks=false]{hyperref}

\usepackage[utf8]{inputenc} 
\usepackage[T1]{fontenc}    
\usepackage{hyperref}       
\usepackage{url}            
\usepackage{booktabs}       
\usepackage{amsfonts}       
\usepackage{nicefrac}       
\usepackage{microtype}      
\usepackage{xcolor}         

\usepackage{graphicx}   
\usepackage{makecell}
\usepackage{color}
\usepackage{amsmath}
\usepackage{amsthm}
\usepackage{hyperref}
\hypersetup{
    colorlinks=true,
    linkcolor=blue,
    filecolor=blue,      
    urlcolor=blue,
    citecolor=cyan,
}
\usepackage{multirow}

\title{DKINet: Medication Recommendation via Domain Knowledge Informed Deep Learning}

\author{%
  Sicen Liu$^1$, Xiaolong Wang$^1$, Xianbing Zhao$^1$, Hao Chen$^2$\thanks{Corresponding author} \\
  $^1$Harbin Institute of Technology (Shenzhen), Shenzhen, China\\
  $^2$The Hong Kong University of Science and Technology, Clear Water Bay, Hong Kong \\
  \texttt{liusicen\_cs@outlook.com,xlwangsz@hit.edu.cn},\\
  \texttt{zhaoxianbing\_hitsz@163.com,jhc@cse.ust.hk} \\
}

\begin{document}

\maketitle


\begin{abstract}
  Medication recommendation is a fundamental yet crucial branch of healthcare that presents opportunities to assist physicians in making more accurate medication prescriptions for patients with complex health conditions. Previous studies have primarily focused on learning patient representation from electronic health records (EHR). While considering the clinical manifestations of the patient is important, incorporating domain-specific prior knowledge is equally significant in diagnosing the patient's health conditions. However, effectively integrating domain knowledge with the patient's clinical manifestations can be challenging, particularly when dealing with complex clinical manifestations. Therefore, in this paper, we first identify comprehensive domain-specific prior knowledge, namely the Unified Medical Language System (UMLS), which is a comprehensive repository of biomedical vocabularies and standards, for knowledge extraction. Subsequently, we propose a knowledge injection module that addresses the effective integration of domain knowledge with complex clinical manifestations, enabling an effective characterization of the health conditions of the patient. Furthermore, considering the significant impact of a patient's medication history on their current medication, we introduce a historical medication-aware patient representation module to capture the longitudinal influence of historical medication information on the representation of current patients. Extensive experiments on three publicly benchmark datasets verify the superiority of our proposed method, which outperformed other methods by a significant margin. The code is available at: https://github.com/sherry6247/DKINet.



\end{abstract}

\section{Introduction}

Combinatorial therapy with the concurrent use of multiple medications is a promising strategy to treat patients with complicated diseases
\cite{almirall2012designing,b3,b5,wu2022leveraging}. Medication recommendation, a fundamental yet crucial branch of healthcare that aims to recommend medications to treat the diagnosed disease of patients, has attracted considerable attention from researchers~\cite{b6,ARMR,MCF}.  Existing approaches primarily focus on exploring the representation patterns of electronic health records (EHR). 
~\cite{leap,MICRON,COGNet}. Part of them tries to utilize the biomedical knowledge from EHR and medication to assist the representation of patients ~\cite{shang2019pre,gao2023medical}. 
However, existing knowledge from EHR and medication suffers limitations in finding effective medication adequately~\cite{yin2019domain,dash2022review}, which deficiency hinders comprehensive insights into the complex health condition of the patient.
Therefore, it is essential to consider external domain knowledge, which encompasses all clinical manifestations (e.g., diagnosis, procedures), to enhance patient representation.

The Unified Medical Language System (UMLS)~\cite{UMLS} is a comprehensive, standardized terminology domain knowledge widely used in healthcare systems. The UMLS merges various subdomain knowledge (e.g., medication, diagnosis, and procedure) in a unified pattern, which is helpful in introducing unified domain knowledge for patients with complex health conditions.  As shown in Figure \ref{figure1}, the diagnosis code \textit{401.9} in the EHR can be mapped to the concept \textit{Essential Hypertension (C0085580)} in the UMLS. The relation knowledge path in UMLS indicates such knowledge can assist in a more comprehensive understanding of the diagnosis and its potential implications for the patient. Therefore, incorporating prior domain knowledge with UMLS information to enhance medication recommendation becomes imperative. 

\begin{figure}[htbp]
  \centering
  \includegraphics[width=0.8\linewidth]{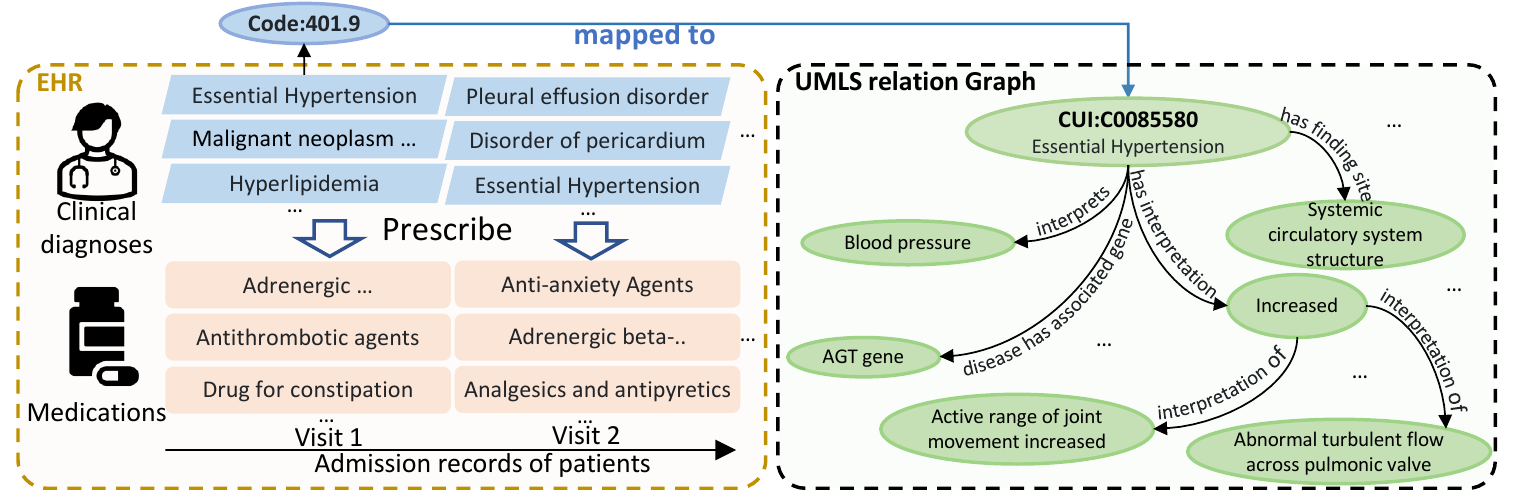}
  \caption{An example mapping of diagnoses and UMLS relation graph. The diagnosis code \textbf{401.9} is mapped to the corresponding CUI code \textbf{C0085580} through the UMLS vocabulary. On the right of the figure, UMLS concepts connected to \textbf{Essential hypertension} are displayed along with their corresponding relationship pathways.}
\end{figure}

Nevertheless, there are several challenges in incorporating relevant domain knowledge aligned with the clinical manifestations to achieve accurate medication recommendations. Firstly, how to retrieve the most related domain knowledge? Secondly, how to gather crucial prior knowledge from a vast number of knowledge graphs? Thirdly, How to efficiently integrate domain knowledge with patients' clinical information to obtain more comprehensive patient representations. Furthermore, the historical medications assume an integral part of the patient assessment in the hospital ~\cite{fitzgerald2009medication}, which is usually ignored in the previous methods~\cite{shang2019pre,sun2022debiased}. 
Hence, a comprehensive medication recommendation method that can effectively combine domain knowledge and historical medication records is strongly needed.

To address these aforementioned challenges, we proposed a novel \textbf{D}omain \textbf{K}nowledge \textbf{I}nformed \textbf{Net}work, dubbed \textbf{DKINet}, to incorporate the UMLS knowledge and observable clinical manifestations. Specifically, we first propose a graph aggregation module to extract the informative knowledge from the UMLS graph effectively. Then, we design a knowledge-injected visit-level representation module to capture the knowledge-enhanced representation of the patient by incorporating the domain knowledge according to the clinical manifestations. Simultaneously, we introduce a regularization term that minimizes mutual information between the knowledge-enhanced and EHR-based representations, enabling our model to absorb external knowledge representation beyond the EHR-based representation.
Moreover, we introduce a historical medication-aware patient representation module to collect historical medication information and capture the longitudinal influence of historical medication on the current patient. Extensive experimental results on three publicly benchmark datasets, \textit{i.e.}, MIMIC-III~\cite{johnson2016mimic}, MIMIC-IV~\cite{johnson2023mimic} and eICU~\cite{pollard2018eicu}, validate the superiority of our proposed method. The main contributions of this article can be summarized as:
\begin{itemize}
    \item We present a new framework to incorporate domain knowledge and observable EHR information toward medication recommendation. To the best of our knowledge, it is the first work to explore the UMLS knowledge for medication recommendation.
    \item We design a graph aggregation module for extracting domain knowledge and a knowledge-injected patient representation module to incorporate domain knowledge based on clinical manifestations.
    We also introduce a historical medication-aware patient representation module for collecting historical medication information.
    \item We evaluated our DKINet on three publicly available benchmark datasets, and it achieved consistently higher performance against the state-of-the-art methods. 
\end{itemize}

\section{Related Works}
\paragraph{Medication Recommendation.}
Medication recommendation algorithms have been developed to assist doctors in making effective medication prescriptions. A series of deep learning based medication recommendation methods have been proposed. For example, Zhang et al.~\cite{leap} decomposed the medication recommendation task into a sequential decision-making process and used the diagnosis of the current visit to generate the medications. Choi et al.~\cite{retain} leveraged the attention module to model the temporal information of the patient. Le et al.~\cite{dmnc} designed a memory mechanism to capture the interaction of asynchronous sequential information. Shang et al.~\cite{gamenet} exploited a graph memory network to model the historical records of the patient for predicting medications. Yang et al.~\cite{MICRON} proposed a residual network to model the changes in medications during the hospitalization of the patient. Yang et al.~\cite{safedrug} utilized the drug molecule structures to enhance the medication encoding representation. Wu et al.~\cite{COGNet} introduced a conditional generation network to generate the medication sequences. Sun et al.~\cite{sun2022debiased} proposed a causal graphical model to deconfound the recommendation bias. Yang et al.~\cite{yangmanydg} treated each patient as a separate domain to develop a many-domain generalization method for medication recommendation.  Although significant progress has been made by considering different patient representation patterns, they are essentially depending on the EHR and overlook the domain knowledge, especially integrating domain knowledge according to the clinical manifestations in the EHR of the patient. To this end, we design a flexible and adaptable domain knowledge informed framework, which could integrate the multi-subdomain knowledge into observable clinical manifestations of the patients.


\paragraph{Knowledge Representation.}
Knowledge graphs (KG) as an auxiliary data source have shown great potential in improving the accuracy and explainability of recommendation tasks~\cite{naumov2019deep,yu2021ernie,yang2021gfe}.  They connect various types of information related to entities in a unified global space, which helps develop insights on recommendation problems that are difficult to uncover with training data only. Researchers usually utilize knowledge information to assist with recommendation tasks~\cite{wang2019multi,cao2019unifying,wang2021learning}.  The UMLS provides a common language for exchanging electronic health information, and it contains domain knowledge that could guide the decision of physicians. Hence, modeling the interaction between UMLS knowledge and the EHR of the patient can help better infer the health condition of the patient for medication recommendation. To the best of our knowledge,  our model is the first one to introduce the domain knowledge informed framework to integrate the UMLS knowledge and observable EHR for medication recommendation.





\section{Methodology}
\subsection{Problem Formulation}
Patient EHR consists of several visit records and the whole data for all patients can be represented as $X=\{X_1, X_2, \cdots, X_N\}$, where $n \in \{1,2, \cdots, N\}$, $N$ is the number of total patients. A specific patient $n$ can be represented as $X_n=[x_n^1, x_n^2, \cdots, x_n^{t}, \cdots]$  and $t \in \{1,2, \cdots, T_n\}$, where $T_n$ is the number of visits for the $n$-th patient. For the $t$-th visit of the $n$-th patient, $x_n^t=[d_n^t, p_n^t,m_n^t]$, where $d_n^t \in \{0,1\}^{\mathcal{D}}, p_n^t \in \{0,1\}^{{\mathcal{P}}}, m_n^t \in \{0,1\}^{{\mathcal{M}}}$ are multi-hot of diagnosis, procedure, and medication vectors, respectively. The $\mathcal{D,P,M}$ are the overall diagnosis, procedure, and medication sets, while $|\cdot|$ denotes the cardinality of the set. The UMLS knowledge includes multi-subdomain concepts and relations between them. It can be denoted as $\mathcal{G}_u= \{\mathcal{H}_u, \mathcal{R}_u, \mathcal{T}_u\}$, where $\mathcal{H}_u$ and $\mathcal{T}_u$ denotes the head and tail entity, respectively. The node set $\mathcal{H}_u + \mathcal{T}_u \gg \mathcal{D} + \mathcal{P} + \mathcal{M}$ contains all concepts in the UMLS, and $\mathcal{R}_u$ contains the relations between the concepts (e.g., \textit{is\_a, mapped\_to, related\_to}). The objective of the medication recommendation task is to recommend medications $\hat{m}_i^t $ at the $t$-th visit of the $i$-th patient, given all the patient's previous visit $[x_i^1, x_i^2, \cdots, x_i^{t-1}]$, current visit input $d_i^t, p_i^t$, and the UMLS graph $\mathcal{G}_u$.

\subsection{The DKINet Framework}
\begin{figure*}[htbp]
	\centering
        \includegraphics[width=\linewidth]{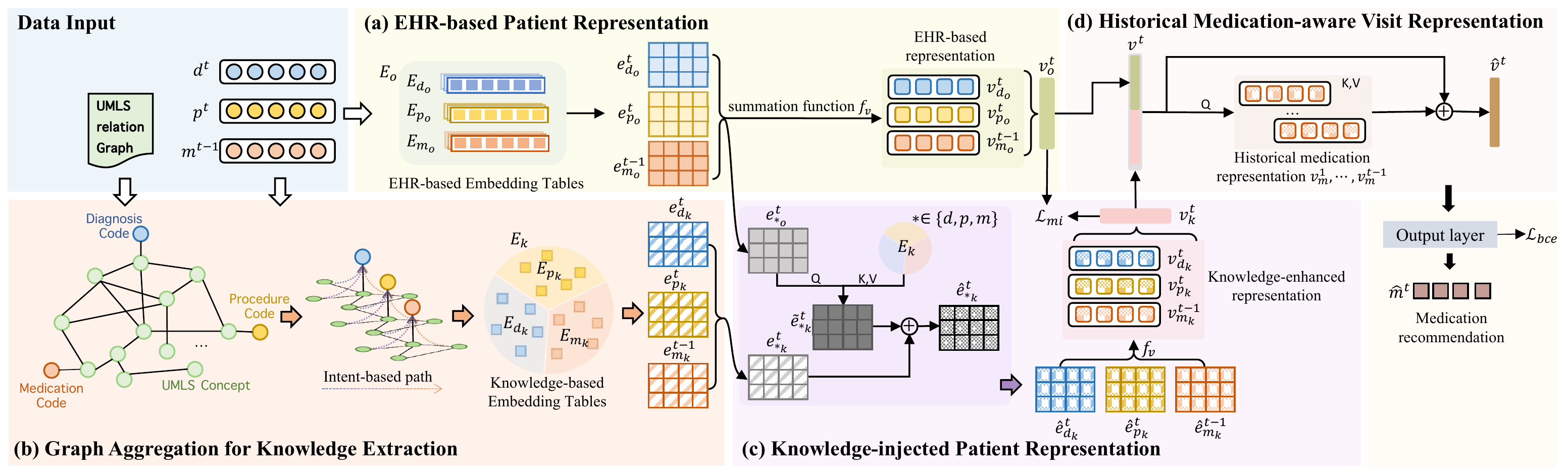}
	\caption{An overview of the proposed DKINet framework. DKINet has four main components: a) EHR-based visit-level representation module takes diagnosis $d^t$, produces $p^t$, and previous visit medication record $m^{t-1}$ as input to get the EHR-based representation of the patient; b) graph aggregation module extracts the domain knowledge information and combine the current EHR input to obtain the knowledge-based embedding vectors $e_{d_k}^t,e_{p_k}^t,e_{m_k}^{t-1}$; c) knowledge-injected visit-level representation module incorporates the domain knowledge with the clinical manifestation and generates the knowledge-enhanced representation; d) historical medication-aware patient representation module is designed to capture the longitudinal influence of historical medication information. Finally, an output layer is designed to make recommendation. }
    \label{figure2}
\end{figure*}

As illustrated in Figure \ref{figure2}, Our proposed DKINet includes the following components: an EHR-based visit-level representation module, a graph aggregation module for knowledge extraction, a knowledge-injected patient representation module, and a historical medication-aware patient representation module. In this section, we elaborate on each component of our DKINet.


\subsubsection{EHR-based Visit-level Representation} The clinical manifestation as the data fundamental to recommending medications reflects the health condition of the patient. The goal of the EHR-based visit-level representation module is to learn the clinical manifestation representation of the patient.
For the $t$-th visit, we use the current diagnosis and procedure record $d^t, p^t$ and the medication record of the previous visit (i.e., the $m^{t-1}$) as input to obtain current clinical manifestations representation. Note that, we apply a padding embedding as the medication input for the first visit of the patient.
Specifically, we design three EHR-based embedding tables $E_o = [E_{d_o}, E_{p_o}, E_{m_o}]$, and convert the three inputs to EHR-based embedding representations $e_{d_o}^t, e_{p_o}^t, e_{m_o}^{t-1}$, respectively. Afterward, we collated the medical code embedding into the visit-level patient representation $v_o^t$:
 \begin{equation}
    v_{*_o}^t = f_v(e_{*_o}^t), * \in \{ d,p,m\}
\end{equation}
\begin{equation}
    v_o^t = [v_{d_o}^t; v_{p_o}^t; v_{m_o}^{t-1}]
\end{equation}
where $f_v$ is a summation function, and $[\cdot;\cdot]$ is the concatenation operation in the feature dimension.
\subsubsection{Graph Aggregation for Knowledge Extraction}
The UMLS graph comprises complex concepts and relationships, and there are limited concepts and relations that are beneficial for specific medical codes.  We hence argue that collecting UMLS knowledge with more critical information is urgently needed. Motivated by this, we design a graph aggregation module, aiming to emulate the act of information-filtering capabilities with physicians as they navigate the vast UMLS knowledge graph. We refer to this new aggregation method as filter-based aggregation, and the corresponding pathway is termed the filter-based path.
Technologically, we first map the medical code $c\in\mathcal{C}$ to the UMLS concept $u \in \mathcal{U}$ according to the vocabulary in UMLS. The symbol $\mathcal{C}$ represents a medical code set $\mathcal{C} =\{\mathcal{D}, \mathcal{P}, \mathcal{M}\} $ and $\mathcal{U} = \{\mathcal{H}_u, \mathcal{T}_u\}$ denotes concepts in the UMLS. The $(c,u)$ denotes the pair of medical code and UMLS concept.
Assuming there are $|\mathcal{F}|$ filters to gather knowledge from the UMLS according to medical code. Then, each pair $(c, u)$ can be transformed into $\{(c, \mathcal{F}_i, u)| \mathcal{F}_{i} \in \mathcal{F}\}$, where $\mathcal{F}_{i}$  embodies the filter meticulously crafted to emulate the knowledge-filtering capabilities of physicians. 
Finally, we obtain sets of filter-based paths between the medical codes and the UMLS concepts and denote them as the filter-based graph $\mathcal{G_F} = \{ \mathcal{C, F, U} \}$, where the node sets are $\{ \mathcal{C, U} \}$ and edge sets are $\{\mathcal{F}\}$ filters.

We decompose the graph aggregation module into two parts: First, we employ the aggregation layer over the UMLS graph to collate the relation graph representation for each concept in the UMLS. Then, we exploit the aggregation layer over the filter-based graph to gather the filter-based information from the UMLS concepts and obtain the knowledge-based representation of medical codes.

\textit{\textbf{Aggregation Layer over UMLS Graph.}} For the UMLS graph, let $\mathcal{N}_{h} = \{(r_h,\tau_h) | (h, r_h, \tau_h) \in \mathcal{G}_u \}$ denoted the sets of the relationships and the neighbor entities about concept $h$, where $h\in\mathcal{U}$. We first gather the knowledge information based on the relation path in UMLS, which is formulated as:
\begin{equation} 
e_{h}^{l_1} = \frac{1}{|\mathcal{N}_h|}\sum_{(r_h, \tau_h) \in \mathcal{N}_h} e_{r_h} \odot e_{\tau_h}^{l_0}
\end{equation}
where $e_{r_h}$ is the relation embedding of $r_h\in \mathcal{R}_u$. The $e_{\tau_h}^{l_0}$ is the initial embedding representation of tail entity $\tau_h$. The $\odot$ is the element-wise product. The $l_1$ donates the first-order connection of concept $h$. Analogously, we can obtain the representation of each concept within the UMLS graph $\mathcal{G}_u$. 

\textit{\textbf{Aggregation Layer over Filter-based Graph.}} The filter-based path aims to gather informative concepts and relations from UMLS. Hence, we assign each filter $\mathcal{F}_i$ with a distribution over KG relations to create the filter embedding as: 
\begin{equation}
    e_{\mathcal{F}} = \sum_{r \in \mathcal{R}_u} \frac{exp(w_{r\mathcal{F}_i})}{\sum_{r' \in \mathcal{R}_u} exp(w_{r'\mathcal{F}_i}) } e_{{\mathcal{R}_u}}
\end{equation}
where $w_{r\mathcal{F}_i} \in \mathbb{R}^{|\mathcal{F}|\times|\mathcal{R}_u|}$ is a trainable weight specific to certain relation $r$ and certain filter $\mathcal{F}_i$. The $e_{\mathcal{R}_u}$ is the relation embedding of UMLS graph $\mathcal{G}_u$.

Let $\mathcal{N}_{c} = \{(\mathcal{F}_i, u) | (c, \mathcal{F}_i, u) \in \mathcal{G}_{\mathcal{F}} \}$ represent the filters and related UMLS concepts of medical code $c$, the aggregation of $c$ is formulated as:
\begin{equation}
    e_{c}^{l_1} = \frac{1}{|\mathcal{N}_{c}|} \sum_{( c, \mathcal{F}_i, u)\in \mathcal{N}_{c}} \frac{exp(e_{\mathcal{F}_i}^{\mathsf{T}} e_c^{l_0})} {\sum_{\mathcal{F}_i' \in \mathcal{F}} exp(e_{\mathcal{F}_i'}^{\mathsf{T}} e_c^{l_0})} e_{\mathcal{F}} \odot e_u^{l_0}
\end{equation}
where $e_c^{l_0}$ and $e_u^{l_0}$ are the initial embedding of medical code and UMLS concept. 
We further stack more aggregation layers to gather knowledge-based information from the higher-order ($l_a$) neighbors. Formally:
\begin{equation}
        e_{h}^{l_a} = \frac{1}{|\mathcal{N}_h|}\sum_{(r_h, t_h) \in \mathcal{N}_h} e_{r_h} \odot e_{t_h}^{l_{a-1}} 
\end{equation}
\begin{equation}
    e_{c}^{l_a} = \frac{1}{|\mathcal{N}_{c}|} \sum_{( c, \mathcal{F}_i, u)\in \mathcal{N}_{c}} \frac{exp(e_{\mathcal{F}_i}^{\mathsf{T}} e_c^{l_{a-1}})} {\sum_{\mathcal{F}_i' \in \mathcal{F}} exp(e_{\mathcal{F}_i'}^{\mathsf{T}} e_c^{l_{a-1}})} e_{\mathcal{F}} \odot e_u^{l_{a-1}}
\end{equation}

To effectively identify the semantics of each filter $\mathcal{F}_i \in \mathcal{F}$, we followed the KGIN~\cite{wang2021learning} to decouple each filter by minimizing the distance correlation of filters.
    \begin{equation}
        \mathcal{L}_{ekg} = \sum_{(\mathcal{F}_i, \mathcal{F}_j) \in \mathcal{F}, i \ne j} \frac{dCov(\mathcal{F}_i, \mathcal{F}_j)}{\sqrt{dVar(\mathcal{F}_i)\cdot dVar(\mathcal{F}_j)}}
    \end{equation}
where $dCov(\cdot)$ is the distance covariance and $dVar(\cdot)$ is the distance variance of each filter representation.

Benefiting from our knowledge graph modeling, these medical code representations can store the UMLS semantics of multi-filters, allowing us to obtain the knowledge-based embedding tables $E_{k} = { E_d^k, E_p^k, E_m^k }$ for all medical codes. Then, given the medical code at $t$-th visit (i.e., $d^t, p^t, m^{t-1}$), we convert the input sequences to knowledge-based embedding representations $e_{d_k}^t, e_{p_k}^t, e_{m_k}^{t-1}$, respectively. 

\subsubsection{Knowledge-injected Patient Representation} 
The graph aggregation module can acquire knowledge-based representation (i.e., $e_{d_k}^t, e_{p_k}^t, e_{m_k}^{t-1}$) directly, but it only captures the local representation of the current code from UMLS. To enhance the acquisition of a comprehensive global representation based on the patient's current clinical manifestation, we introduce the knowledge-injected patient representation module to seamlessly incorporate the domain knowledge according to the clinical manifestations.
Concretely, we use the EHR-based medical code representation as the query to acquire the global knowledge-based representation from the knowledge-base embedding table $E_k$ and concatenate it with the local knowledge-based representation:
\begin{equation}
    \hat{e}_{*_k}^t = softmax({e_{*_o}^t}^{\mathsf{T}} {E_k} ) E_k , * \in \{ d,p,m\}
\end{equation}
\begin{equation}
    v_{*_k}^t = f_v([e_{*_k}^t;\hat{e}_{*_k}^t]), * \in \{ d,p,m\}
\end{equation}
where $e_{*_o}^t $ is the EHR-based medical code embedding vectors and $\hat{e}_{*_k}^t $ is the knowledge-enhanced medical code embedding vector. The $f_v$ is a summation function and $[\cdot;\cdot]$ is the concatenation operation. Eventually, we concatenate the three types of medical code embedding to obtain the knowledge-enhanced patient representation $v_k^t$ as follows:
\begin{equation}
    v_k^t = [v_{d_k}^t; v_{p_k}^t; v_{m_k}^{t-1}]
\end{equation}

After acquiring the knowledge-enhanced representation, we concatenate it with the EHR-based feature to form the current visit-level representation of the patient:
\begin{equation}
    v^t = w_v[v_k^t; v_o^t]+b_v
\end{equation}
where $w_v \in \mathbb{R}^{2dim \times dim}$ and $b_v \in \mathbb{R}^{dim}$ are learnable parameters and $dim$ denotes the dimension of the feature, and $[\cdot;\cdot]$ denotes concatenation operation in the feature dimension.

\textit{\textbf{Mutual Information Minimization Regularization.}}
To further capture the domain knowledge and enrich the representations of the patient, we design an additional enhancement constraint to regularize the enhanced part of the representation.
Inspired by CLUB~\cite{cheng2020club}, we utilize a regularization term grounded in mutual information minimization to amass broader domain knowledge.
Concretely, with sample knowledge-enhanced and EHR-based representation pairs $\{(v_o^s, v_k^s)\}_{s=1}^\mathcal{S}$, we minimize the distance of latent space between the $v_k^s$ and $v_o^s$ by optimizing mutual information (MI) upper bounds as follows:
\begin{equation}
    \mathcal{L}_{mi} 
    = \frac{1}{\mathcal{S}^2}\sum_{s=1}^\mathcal{S} \sum_{r=1}^\mathcal{S} [\log f_{\theta}(v_k^s|v_o^s)-\log f_{\theta}(v_k^r|v_o^s)]
\end{equation}
where $f_{\theta}(\cdot)$ is a variation approximation function and $f_{\theta}(.)$ is parameterized by:
\begin{equation}
    f_{\theta}(v_k^i|v_o^i) = \mathcal{N}(\mu_{\theta}(v_o^i), {I})
\end{equation}
where $\mathcal{N}(\cdot)$ is Gaussian distribution.

\subsubsection{Historical medication-aware patient Representation} 
The medication historical records in the EHR  of patients are crucial for preventing prescription errors and consequent risks to patients~\cite{fitzgerald2009medication}. 
To capture the longitudinal dependence of historical medication on the current visit representation, we adopt a historical medication-aware patient representation module to model this relationship effectively. In particular, we use current patient representation $v^t$ as the query, the historical medication representation sequences $(v_m^{1},\cdots, v_m^{t-1})$ as the key and value to capture dependency of historical medication information,
\begin{equation}
    \hat{v}^t = softmax(\frac{{v^t}^{\mathsf{T}} (v_m^{1},\cdots, v_m^{t-1})}{\sqrt{dim}})(v_m^{1},\cdots, v_m^{t-1}) + v^t
\end{equation}

Finally, we adopt an output layer to obtain the final medication representation $\hat{y}^t$ as follows:
\begin{equation}
    \hat{y}^t = \sigma(w_y \hat{v}^t + b_y)
\end{equation}
where $w_y \in \mathbb{R}^{dim \times |\mathcal{M}|}$ and $b_y \in \mathbb{R}^{|\mathcal{M}|}$ are learnable parameters, and $\sigma$ is sigmoid function.


\subsection{Objective Function}
Following the previous works~\cite{safedrug,sun2022debiased}, we treat the medication recommendation task as a multi-label prediction task and adopt the binary cross entropy loss $\mathcal{L}_{bce}$ 
as loss function.
\begin{equation}
    \mathcal{L}_{bce} = -\sum_{t}^{T}\sum_{z}^{|\mathcal{M}|} m_z^{t}log(\hat{y}_z^{t}) + (1 - m_z^{t})log(1 - \hat{y}_z^{t})
\end{equation}
where $m_z^t$ is the ground truth medication code at $z$-th coordinate at $t$-th visit, and the $\hat{y}_z^t$ is the probability of the predicted medication code. The overall learning objective of {DKINet} is performed by minimizing the combined loss function:
\begin{equation}
    \mathcal{L} = \mathcal{L}_{bce} + \alpha \mathcal{L}_{ekg} + \beta \mathcal{L}_{mi} 
\end{equation}
where $\alpha, \beta$ are hyperparameters. During the inference phase, the predicted medications are generated from the $\hat{y}^t$. a threshold $\eta$ is used to distinguish the correctly predicted labels. Then, the predicted label set corresponds to the following:
\begin{equation}
    \hat{m}^t = \{\hat{y}_z^t|\hat{y}_z^t \ge \eta, 1\le z \le |\mathcal{M}|\}
\end{equation}
\begin{equation}
    \mathcal{L}_{bce} = -\sum_{t}^{T}\sum_{i}^{|\mathcal{M}|} m_i^{t}log(\hat{y}_i^{t}) + (1 - m_i^{t})log(1 - \hat{y}_i^{t})
\end{equation}

 where $m_i^t$ is the ground truth medication code at $i$-th coordinate at $t$-th visit, and the $\hat{y}_t^t$ is the probability of the predicted medication code corresponding to $m_i^t$. The overall learning objective of {DKINet} is performed by minimizing the combined loss function:
\begin{equation}
    \mathcal{L} = \mathcal{L}_{bce} + \alpha \mathcal{L}_{ekg} + \beta \mathcal{L}_{mi} 
\end{equation}
where $\alpha, \beta$ is hyperparameter. During the inference phase, the predicted medications are generated from the $\hat{y}^t$. We use threshold $\eta$ to distinguish the correctly predicted labels. Then, the predicted label set corresponds to the following:
\begin{equation}
    \hat{m}^t = \{\hat{y}_i^t|\hat{y}_i^t \ge \eta, 1\le i \le |\mathcal{M}|\}
\end{equation}

\section{Experiments}
\subsection{Experimental Setup}
\subsubsection{Datasets and Pre-processing}

\begin{table*}[]
    \centering    
    \scriptsize
    \caption{The statistics of the eventual datasets and UMLS}
    \begin{tabular}{c|c|c|c}
    \hline
        Item & MIMIC-III & MIMIC-IV & eICU\\
        \hline
        \# of patients & 5,208 & 6,136 & 9,583\\
         \# of clinical visits & 12,490 & 17,813 & 21,018 \\
        \# of diag. & 1,895 & 1,851 & 1,371 \\
        \# of proc. & 1,378 & 4,001 & - \\
        \# med.  & 112 & 121 & 45 \\
        avg. \# of visits & 2.59 & 2.90 & 2.24  \\
        avg. \# of diag. / vi. & 10.24 & 11.78 & 8.06 \\
        avg. \# of proc. / vi. & 3.85 & 2.18 & - \\
        avg. \# of med. / vi. & 11.30 & 6.68 & 9.99 \\
        \hline
        \# concept & 112,298 & 101,091 & 158,340 \\
        \# semantic type  & 28 & 31 &  16 \\
        \# relationship & 287 & 289 &  221  \\
    \hline
    \end{tabular}
    \label{table1}
\end{table*}

To justify the effectiveness of our model, we conducted experiments under publicly available MIMIC-III~\cite{johnson2016mimic}, MIMIC-IV~\cite{johnson2023mimic}, and eICU~\cite{pollard2018eicu} datasets. We follow the data processing of DrugRec~\cite{sun2022debiased} that removes those patients who only have a single visit record.
For MIMIC-III/IV datasets, the ICD codes are used for diagnosis and procedure while mapping NDC codes to the third-level ATC codes for labeling medication. For the eICU dataset, we collated the ICD code of diagnosis and GTC code of medication due to the eICU dataset itself lacking the procedure records. 
We used the Unified Medical Language System (UMLS)\footnote{https://www.nlm.nih.gov/research/umls/index.html} as external knowledge.
There are three sources of UMLS: \textit{Metathesaurus, Semantic Network}, and \textit{SPECIALIST Lexicon} \& \textit{Lexical Tools}. 
Following the UMLSParse\footnote{https://github.com/DATEXIS/UMLSParser},  we extracted the unified CUI code of concepts from "MRCONSO.RRF", "SERDF" and "SRSTRE1" file in \textit{Metathesaurus} folder and collated the categories of semantic type from "SRDEF" file in \textit{Semantic Network} folder. For the relationships, we use the relationship that appears in the "MRREL.RRF" and "MRSTY.RRF" files.
The medical codes are mapped to the CUI codes according to the "MRCONSO.RRF" file in \textit{Metathesaurus} folder.
The statistical information of the three datasets is listed in Table \ref{table1}.

\begin{table*}[]
    \centering
    \caption{Experimental results on the MIMIC-III datasets.}
    \begin{tabular}{l|ccccc}
    \hline 
      \textbf{KG} & \textbf{Methods} & \textbf{Jaccard\textsubscript{(std.)}}  & \textbf{F1\textsubscript{(std.)}} & \textbf{PRAUC\textsubscript{(std.)}} & \textbf{DDI\textsubscript{(std.)}}  \\
      \hline 
      - &LR & 0.4896\textsubscript{0.0025} & 0.6491\textsubscript{0.0024}  & 0.7568\textsubscript{0.0025 } & 0.0774\textsubscript{.0012}    \\
      - &ECC~\cite{read2011classifier} & 0.4799\textsubscript{0.0022} & 0.6390\textsubscript{0.0022}  &  0.7572\textsubscript{0.0026} & 0.0760\textsubscript{0.0010} \\
      - &RETAIN~\cite{retain} & 0.4780\textsubscript{0.0036} & 0.6397\textsubscript{0.0036}  & 0.7601\textsubscript{0.0035} & 0.0814\textsubscript{0.0018}  \\
      - &LEAP~\cite{leap}(2017) & 0.4465\textsubscript{0.0037} & 0.6097\textsubscript{0.0036} & 0.6490\textsubscript{0.0033} & 0.0657\textsubscript{0.0010} \\
      - &MICRON~\cite{MICRON} & 0.5076\textsubscript{0.0037} & 0.6634\textsubscript{0.0035} & 0.7685\textsubscript{0.0038} & 0.0612\textsubscript{0.0008} \\
      - &ManyDG~\cite{yangmanydg} & 0.4090\textsubscript{0.0046} & 0.5745\textsubscript{0.0031} & 0.6891\textsubscript{0.0020} & 0.0798\textsubscript{0.0023} \\
      \hline 
      \checkmark &G-BERT~\cite{shang2019pre} & 0.4651\textsubscript{0.0016}  & 0.6267\textsubscript{0.0015} & 0.7510\textsubscript{0.0017} & 0.0811\textsubscript{0.0011} \\
      \checkmark &GAMENet~\cite{gamenet}  & 0.5039\textsubscript{0.0021}  & 0.6609\textsubscript{0.0020} & 0.7632\textsubscript{0.0027}  & 0.0832\textsubscript{0.0005} \\
      \checkmark &SafeDrug~\cite{safedrug} & 0.5090\textsubscript{0.0038} & 0.6664\textsubscript{0.0033} &0.7647\textsubscript{0.0020}  & 0.0658\textsubscript{0.0003} \\
      \checkmark &COGNet~\cite{COGNet} & 0.5134\textsubscript{0.0027} & 0.6706\textsubscript{0.0043} & 0.7677\textsubscript{0.0013}  & 0.0784\textsubscript{0.0005}  \\ 
      \checkmark &DrugRec~\cite{sun2022debiased} & 0.5220\textsubscript{0.0034} & 0.6771\textsubscript{0.0031} & 0.7720\textsubscript{0.0036}  & 0.0597\textsubscript{0.0006} \\ 
      \checkmark &MK-GNN~\cite{gao2023medical} & \underline{0.5238\textsubscript{0.0015}} & \underline{0.6775\textsubscript{0.0014}} & \underline{0.7746\textsubscript{0.0014}} & \underline{0.0623\textsubscript{0.0012}}\\
      \midrule
      \checkmark &\textbf{DKINet} & \textbf{0.5345\textsubscript{0.0013} } & \textbf{0.6880\textsubscript{0.0012} } & \textbf{0.7896\textsubscript{0.0010}}  & 0.0600\textsubscript{0.0003} \\
    \hline 
    \end{tabular}
    \label{table2}
\end{table*}

\begin{table*}[]
    \centering
    \caption{Experimental results on the MIMIC-IV datasets.}
    \begin{tabular}{l|ccccc}
    \hline
          \textbf{KG} & \textbf{Methods} & \textbf{Jaccard\textsubscript{std.}}  & \textbf{F1\textsubscript{std.}} & \textbf{PRAUC\textsubscript{std.}}  & \textbf{DDI\textsubscript{std.}}  \\
          \midrule
          - &LR &  0.3844\textsubscript{0.0028}  & 0.5379\textsubscript{0.0031} & 0.6568\textsubscript{0.0036} & 0.0645\textsubscript{0.0012} \\
          - &ECC~\cite{read2011classifier} & 0.3690\textsubscript{0.0041} & 0.5173\textsubscript{0.0047} &  0.6541\textsubscript{0.0030} & 0.0648\textsubscript{0.0018} \\
          - &RETAIN~\cite{retain} & 0.3903\textsubscript{0.0038}  & 0.5471\textsubscript{0.0040}  & 0.6563\textsubscript{0.0055}   & 0.0618\textsubscript{0.0025}  \\
          - &LEAP~\cite{leap} & 0.3653\textsubscript{0.0028} & 0.5201\textsubscript{0.0033} & 0.5314\textsubscript{0.0038}  & 0.0570\textsubscript{0.0011} \\          
          - &MICRON~\cite{MICRON}  & 0.4009\textsubscript{0.0044}  & 0.5545\textsubscript{0.0048}  & 0.6584\textsubscript{0.0043}   & 0.0605\textsubscript{0.0017} \\
          - &ManyDG~\cite{yangmanydg} & 0.3146\textsubscript{0.0052} & 0.4718\textsubscript{0.0049} & 0.5816\textsubscript{0.0007} & 0.0712\textsubscript{0.0013} \\
          \midrule
          \checkmark &G-BERT~\cite{shang2019pre} & 0.3693\textsubscript{0.0016} & 0.5236\textsubscript{0.0019} & 0.6266\textsubscript{0.0021} & 0.0811\textsubscript{0.0011} \\
          \checkmark &GAMENet~\cite{gamenet} & 0.3957\textsubscript{0.0035} & 0.5525\textsubscript{0.0041}  & 0.6479\textsubscript{0.0055}  & 0.0757\textsubscript{0.0014}  \\
          \checkmark &SafeDrug~\cite{safedrug} & 0.4082\textsubscript{0.0026}  & 0.5651\textsubscript{0.0028}  & 0.6495\textsubscript{0.0036}   & 0.0553\textsubscript{0.0010}  \\
          \checkmark &COGNet~\cite{COGNet} & 0.4131\textsubscript{0.0020}  & 0.5660\textsubscript{0.0019}  & 0.6460\textsubscript{0.0017}  & 0.0596\textsubscript{0.0005}  \\
          \checkmark &DrugRec~\cite{sun2022debiased} & 0.4194\textsubscript{0.0020}  & 0.5713\textsubscript{0.0022}  & 0.6558\textsubscript{0.0026}  & \underline{0.0396 \textsubscript{0.0007}} \\
          \checkmark &MK-GNN~\cite{gao2023medical} & \underline{0.4234\textsubscript{0.0015}} & \underline{0.5793\textsubscript{0.0016}} & \underline{0.6753\textsubscript{0.0019}} & 0.0566\textsubscript{0.0004}\\
          \midrule
          \checkmark &\textbf{DKINet} & \textbf{0.4420\textsubscript{0.0016}} & \textbf{0.5966\textsubscript{0.0018}} & \textbf{0.6884\textsubscript{0.0016}}  & 0.0404\textsubscript{0.0002}  \\ 
    \hline
    \end{tabular}
    \label{table3}
\end{table*}

\begin{table*}[htbp]
    \centering
    \caption{Experimental results on the eICU datasets.}
    \begin{tabular}{l|cccc}
    \hline
          \textbf{KG} & \textbf{Methods} & \textbf{Jaccard\textsubscript{std.}}  & \textbf{F1\textsubscript{std.}} & \textbf{PRAUC\textsubscript{std.}}  \\
          \midrule
          - &LR & 0.4026\textsubscript{0.0016} & 0.5548\textsubscript{0.0016} & 0.7364\textsubscript{0.0028} \\
          - &ECC~\cite{read2011classifier} & 0.3289\textsubscript{0.0049} & 0.4634\textsubscript{0.0056} & 0.7334\textsubscript{0.0027} \\
          - &RETAIN~\cite{retain} & 0.4118\textsubscript{0.0037} & 0.5590\textsubscript{0.0036} & 0.6992\textsubscript{0.0049} \\
          - &LEAP~\cite{leap} & 0.3033\textsubscript{0.0048} & 0.4358\textsubscript{0.0060} & 0.4826\textsubscript{0.0040} \\
          - &MICRON~\cite{MICRON} & 0.3703\textsubscript{0.0021} & 0.5174\textsubscript{0.0024} & 0.7346\textsubscript{0.0031} \\
          - &ManyDG~\cite{yangmanydg}& 0.4077\textsubscript{0.0051} & 0.5603\textsubscript{0.0030} & 0.7042\textsubscript{0.0023} \\
          \midrule
          \checkmark &G-BERT~\cite{shang2019pre}  & 0.3937\textsubscript{0.0021} & 0.5390\textsubscript{0.0022} & 0.6726\textsubscript{0.0024} \\
          \checkmark &GAMENet~\cite{gamenet} & \underline{0.4519\textsubscript{0.0023}} & \underline{0.6008\textsubscript{0.0026}} & 0.7371\textsubscript{0.0031} \\
          \checkmark &SafeDrug~\cite{safedrug} & 0.4422\textsubscript{0.0031} & 0.5903\textsubscript{0.0035} & 0.7378\textsubscript{0.0042}\\
          \checkmark &COGNet~\cite{COGNet} & 0.4486\textsubscript{0.0014} & 0.5930\textsubscript{0.0014} & 0.7160\textsubscript{0.0018} \\
          \checkmark &DrugRec~\cite{sun2022debiased} & 0.4396\textsubscript{0.0022} & 0.5882\textsubscript{0.0022} & 0.7286\textsubscript{0.0023} \\
          \checkmark &MK-GNN~\cite{gao2023medical} &  0.4451\textsubscript{0.0013} & 0.5936\textsubscript{0.0013} & \underline{0.7467\textsubscript{0.0013}} \\
          \hline
          \checkmark &\textbf{DKINet} &  \textbf{0.4759\textsubscript{0.0016}} & \textbf{0.6209\textsubscript{0.0018}} & \textbf{0.7507\textsubscript{0.0018}} \\ 
    \hline
    \end{tabular}
    \label{table4}
\end{table*}

\noindent\textbf{Implementation details.}
We follow the setting of DrugRec~\cite{sun2022debiased}, dividing the data into training, validation, and test sets with a ratio of $4:1:1$ for each dataset.  Our method was implemented by PyTorch 1.7.1 based on Python 3.8.13, training on GeForce RTX 3090 GPUs. Models are trained on Adam~\cite{kingma2014adam} optimizer with the learning rate of 1e-3 and batch size of 4. We choose the optimal parameters based on the validation set, where the feature dimension size is 256, the number of filters is 4, the knowledge aggregation layers are 1, and the threshold of output $\eta$ is 0.5, the $\alpha$ and $\beta$ of $\mathcal{L}_{ekg}$ and $\mathcal{L}_{mi}$ are set as 1.

\subsubsection{Compared Methods and Metrics}
To justify the effectiveness of our proposed model, we compared it with several state-of-the-art baselines and classified them into two categories based on whether they used external knowledge.

For the methods that do not employ external knowledge,
    \textbf{LR} is standard logistic regression.
    \textbf{ECC}~\cite{read2011classifier} uses multiple SVM~\cite{hearst1998support} classifiers to make predictions.
    \textbf{RETAIN}~\cite{retain} employs an attention module to collate temporal information.
    \textbf{LEAP}~\cite{leap} leverages the LSTM-based generation method to recommend the medication based on the current diagnosis.
    \textbf{MICRON}~\cite{MICRON} focuses on the differences between medications and utilizes a residual-based network to update patient features.
    \textbf{ManyDG}~\cite{yangmanydg} treated each patient as a separate domain and developed a many-domain generalization method to predict medications.

Among the methods that incorporate external knowledge,
    \textbf{G-BERT}~\cite{shang2019pre} integrates the diagnosis and medication ontology hierarchical structures into a transformer-based visit encoder.
    \textbf{GAMNet}~\cite{gamenet} incorporates the drug-drug interaction graph and medication co-occurrence graph using a memory module.
    \textbf{safeDrug}~\cite{safedrug} employs drug molecule structures to enhance the representation of medications, leading to improved medication prediction.
    \textbf{COGNet}~\cite{COGNet} combines the drug-drug interaction graph and generates medications step-by-step.
    \textbf{DrugRec}~\cite{sun2022debiased} considers recommendation bias as a confounding factor and employs a causal graph approach.
    \textbf{MK-GNN}~\cite{gao2023medical} integrates the prior knowledge from EHR and medication knowledge to medication recommendation.

To measure the prediction performance of our proposed DKINet method, we follow the previous works~\cite{sun2022debiased,gao2023medical}, using Jaccard Similarity Score (Jaccard), Average F1 (F1). Precision-Recall AUC (PRAUC) as the evaluation metrics. We also compared the DDI values of different methods on both MIMIC-III and MIMIC-IV for reference.

\subsection{Quantitative Evaluation and Comparison} 
\subsubsection{Result Analysis}
Following the previous work~\cite{sun2022debiased}, the results in Table \ref{table2} were obtained through 10 rounds of bootstrapping sampling on the test set. 
By analyzing these results, we gained the following observations: 
These methods that do not explore external knowledge, such as RETAIN, and MICRON, outperform LR, ECC, and LEAP. Because they explore the longitudinal representation pattern of the patient.
For these methods utilized external knowledge, G-BERT is pre-trained based on single visit records so it exhibited a decrease in performance when transferred to multi-visit and new datasets.
GAMENet and SafeDrug improved performance by incorporating medication-related knowledge. These results reveal that introducing external knowledge plays a meaningful role in medication recommendation.
The performance of COGNet indicates the importance of aggregating the historical medication records. The result of DrugRec verifies the crucial importance of clinical manifestations in medication recommendation.
The result of MK-GNN reveals that jointly modeling the clinical manifestation contributes to a more powerful representation of the patient.
Our proposed DKINet outperformed the compared methods and achieved the best Jaccard, F1, and PRAUC. Compared with the MK-GNN method, DKINet achieves improvements with nearly 1.07\% Jaccard gain, 1.05\% F1 gain, and a relative 1.50\% PRAUC gain on the MIMIC-III. In the MIMIC-IV datasets, our DKINet achieves 1.86\% Jaccard gain 1.73\% F1 gain, and  1.31\% PRAUC gain compare to the MK-GNN method. On the eICU dataset, compared with the GAMENet method, our DKINet obtained a relative 2.40\% Jaccard and 2.01\% F1 gain. The improvement indicates the significance of exploring richer domain knowledge by the clinical manifestations. 



\begin{table}[]
    \centering
    \caption{Ablation study of DKINet on the MIMIC-III, MIMIC-IV, and eICU datasets.}
    \scriptsize
    \begin{tabular}{ccccccccc}
    \hline %
    \multirow{1}{*}{Variants} & \multirow{1}{*}{Dataset} & Jaccard  & F1 & PRAUC ($p$-value) \\
    \hline 
    \textbf{DKINet}  & \textbf{MIMIC-III} &  \textbf{0.5345} & \textbf{0.6880} & \textbf{0.7896}  \\
    DKINet$_{w/o KG}$ & MIMIC-III & 0.5078  & 0.6644  & 0.7790 (7e-14) \\
    DKINet$_{w/o FKG}$ & MIMIC-III & 0.5244 & 0.6794 & 0.7848 (3e-9) \\
    DKINet$_{w/o KI}$ & MIMIC-III & 0.5216  & 0.6772 & 0.7816 (2e-11) \\
    DKINet$_{w/o HMA}$ & MIMIC-III & 0.5280  & 0.6822 & 0.7838 (6e-11) \\
    \hline 
    \textbf{DKINet} & \textbf{MIMIC-IV} & \textbf{0.4420} & \textbf{0.5966} & \textbf{0.6884} \\
    DKINet$_{w/o KG}$ & MIMIC-IV & 0.4152 & 0.5690 & 0.6729 (4e-14) \\
    DKINet$_{w/o FKG}$ & MIMIC-IV &  0.4214 & 0.5756 & 0.6813 (2e-7) \\
    DKINet$_{w/o KI}$ & MIMIC-IV & 0.4183 & 0.5725 & 0.6833 (6e-6) \\
    DKINet$_{w/o HMA}$ & MIMIC-IV & 0.4200 & 0.5745 & 0.6796 (4e-8)  \\
    \hline 
    \textbf{DKINet}  & \textbf{eICU} & \textbf{0.4759} & \textbf{0.6209} & \textbf{0.7507} \\
    DKINet$_{w/o KG}$ & eICU &  0.4249 & 0.5753  & 0.7440 (3e-8) \\
    DKINet$_{w/o FKG}$ & eICU &  0.4351  & 0.5862 & 0.7456 (4e-5) \\
    DKINet$_{w/o KI}$ & eICU & 0.4361 & 0.5867 & 0.7409 (3e-11) \\
    DKINet$_{w/o HMA}$ & eICU & 0.4317 & 0.5811  & 0.7433 (2e-7) \\
    \hline 
    \end{tabular}
    \label{table5}
\end{table}

\begin{figure}[]
    \centering
     \includegraphics[width=\linewidth]{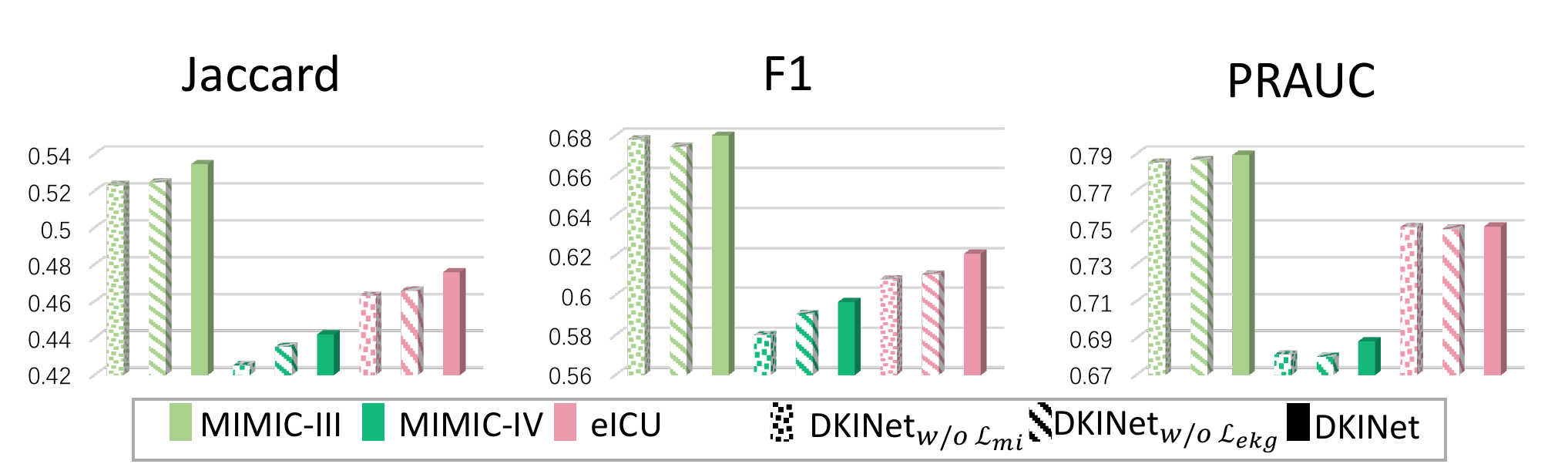}
    \caption{Performance comparison of DKINet, DKINet$_{w/o \mathcal{L}_{mi}}$ and DKINet$_{w/o \mathcal{L}_{ekg}}$ on three metrics (i.e., Jaccard, F1, PRAUC), across the MIMIC-III, MIMIC-IV, and eICU datasets.}
    \label{figure_loss}
\end{figure}


\subsubsection{Ablation Study}
To validate the impact of each module in our proposed DKINet, we conducted ablation studies incrementally. 
Specifically, we compared DKINet with the following variants:
1) DKINet$_{w/o KG}$, removing the UMLS graph;
2) DKINet$_{w/o FKG}$, without the aggregation layer over filter-based graph; 
3)DKINet$_{w/o KI}$, eliminating the knowledge-injected module; 
4) DKINet$_{w/o HMA}$, replacing the attention in the historical medication-aware patient representation module with a self-attention mechanism. 
As reported in Table \ref{table5},
the performance of DKINet$_{w/o KG}$ degrades dramatically, which demonstrates the vital importance of domain knowledge.
Besides, DKINet$_{w/o FKG}$ achieves better results than DKINet$_{w/o KG}$, revealing the effectiveness of the aggregation layer over the filter-based graph to collate domain knowledge and boost the model performance.
The performance drop of DKINet$_{w/o KI}$ indicates that it is vital to consider the local and global aspects to incorporate the domain knowledge. 
Moreover, the performance of DKINet$_{w/o HMA}$ emphasizes the critical role of historical medication information.
We also conducted a significant test between DKINet and the variety of DKINet on the PRAUC metric, and the results demonstrate the statistical significance of our model DKINet. 
In summary, our proposed DKINet surpasses all variant methods, validating the complementarity of each module.
To further verify the impact of we proposed loss function $\mathcal{L}_{mi}$ and $\mathcal{L}_{ekg}$, we illustrated the performance of DKINet, DKINet$_{w/o \mathcal{L}_{mi}}$ and DKINet$_{w/o \mathcal{L}_{ekg}}$ on three datasets, as shown in Figure \ref{figure_loss}. Similar to previous variants of DKINet, DKINet$_{w/o \mathcal{L}_{mi}}$ and DKINet$_{w/o \mathcal{L}_{ekg}}$ mean remove the loss function $\mathcal{L}_{mi}$ and $\mathcal{L}_{ekg}$, respectively. This stable performance across the three datasets may indicate the loss functions are capable of enabling the model to explore the more comprehensive patient representation.

\subsection{Discussion}
\subsubsection{Filter-based Aggregation Analysis}

\begin{figure}[]
    \centering
     \includegraphics[width=\linewidth]{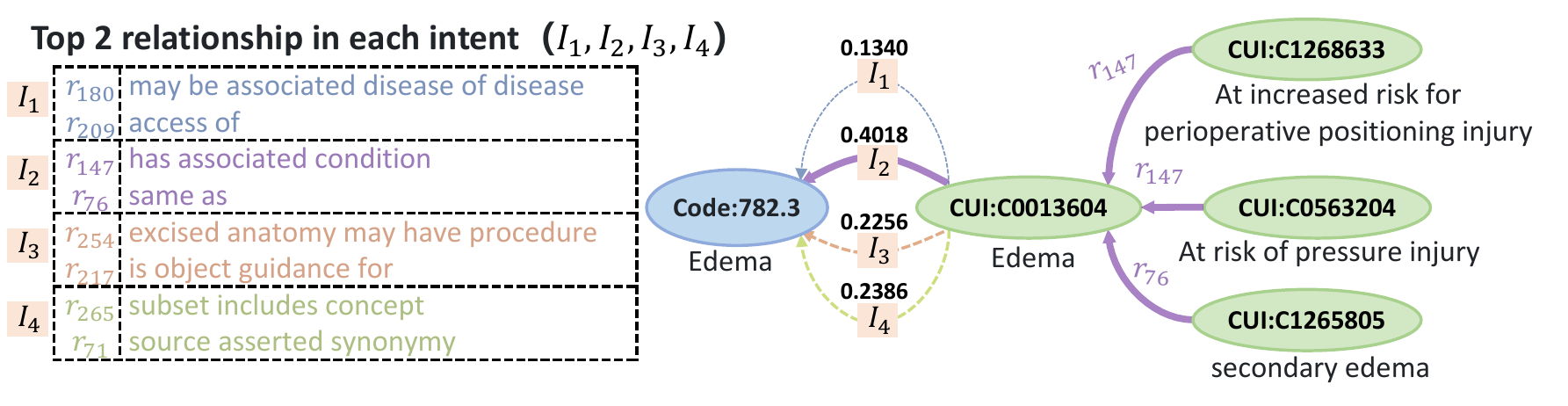}
    \caption{Explanations of filter-base path aggregation and real example in MIMIC-III dataset. Best viewed in color.}
    \label{figure5}
\end{figure}

To gain insights into our filter-based information filtering capabilities, we present an example of the MIMIC-III dataset to provide an intuitive impression of our filter-based aggregation. As shown in Figure \ref{figure5}, our filter-base aggregation initially induces the filters ($F_1, F_2, F_3, F_4$) with various combinations of KG relations. The learned filters abstract the reason for knowledge filtering. We also exhibit the top 2 relationships for each filter, where the weight of a relation reflects its significance in filtering the UMLS knowledge. In detail, for the code \textit{782.3}, DKINet searches the most influential filter $F_2$ based on Eq. (2) and gathers the UMLS knowledge based on the various relations (e.g., \textit{has associated condition}). These observations indicate that filter-based aggregation can effectively gather UMLS knowledge by emulating the act of information-filtering capabilities with physicians, thereby retrieving the most related domain knowledge.


\subsubsection{Knowledge Injection Analysis}
\begin{figure}[]
    \centering
     \includegraphics[width=\linewidth]{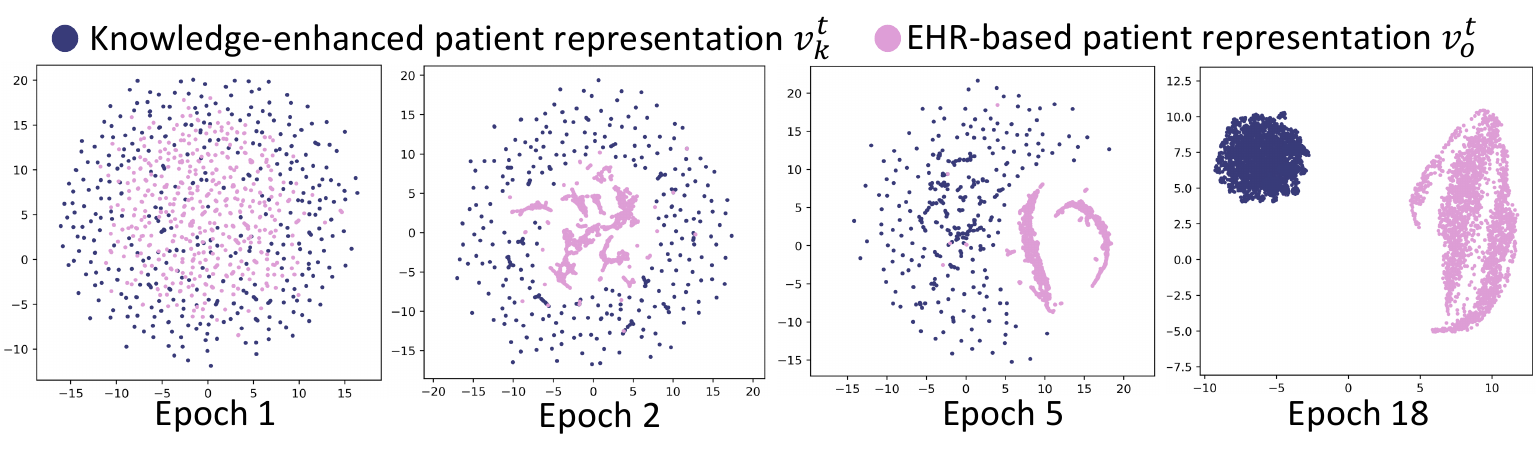}
    \caption{A visualization of the distribution between knowledge-enhanced and EHR-based visit representation in MIMIC-III dataset at different training epochs employing Umap.}
    \label{figure4}
\end{figure}

To justify the effectiveness of knowledge injection, we employed Umap~\cite{mcinnes2018umap} tool to visualize the distribution of EHR-based and knowledge-enhanced representation during various training epochs on the MIMIC-III dataset. As shown in Figure \ref{figure4}, we could find that as the training progresses, it becomes evident that the disparity between the two representations gradually expands. The result demonstrates that the domain knowledge we acquire is independent of EHR-based representation and serves to enhance EHR-based representations.
\subsubsection{Case Study}


We use a case study on the MIMIC-IV dataset to show the ability of domain knowledge on medication recommendation.
Tabel \ref{table6} displays the result of medication recommentation of DKINet and DKINet$_{w/o KG}$. 
Compared to the DKINet$_{w/o KG}$,
DKINet achieves 11 and 12 correct medication recommendations for two visits, with only 3 and 4 missed medications, and wrongly predicts 1 (unseen) medication for the 2nd visit. Besides, the medication predictions of DKINet$_{w/o KG}$ encompass fewer accurate results and exhibit a higher rate of erroneous predictions. 
Furthermore, we noticed that the DKINet can capture more medication candidates and decrease missed results. This phenomenon may unveil the impact of incorporating domain knowledge to improve the accuracy of medication recommendations.

\begin{table}[]
    \centering
    \caption{A case of recommended medications for a given patient's health condition on MIMIC-IV dataset.}
    \scriptsize
    \begin{tabular}{c|c|c|c|c} 
        \hline
        \textbf{Case  } & \multicolumn{4}{c}{\textbf{Code}}  \\
        \hline
        1st Diagnoses: & \multicolumn{4}{l}{\makecell{E872,K2970,Z8673,F329,I471,F17210,D696,R509,Z9114,\\G8929,Z8619,B20,R51,F10239,R197,I959,  G92,G40909,$\cdots$}} \\
        \hline
        1st Procedure:  & \multicolumn{4}{l}{\makecell{009U3ZX} }\\
        \hline
        1st Target: &\multicolumn{4}{l}{\makecell{A11D,J05A,N05A,N05B,J01D,A02A,A12A,C07A,A07A,\\A01A,B05C,A06A,A12C,B01A (14 codes)}} \\
        \hline
        Model & Predicted codes & \textbf{hit} & missed & {unseen} \\
        \hline
        DKINet$_{w/o KG}$ & \makecell{\textbf{B01A}, {J01E},\textbf{A06A}, {N02B},\\ \textbf{A12C},\textbf{A01A}, {N02A}, \textbf{N05B}, \\ {N06A},\textbf{N05A},\textbf{A02A},\textbf{A12A}, \\ {N03A}} & \textbf{8} & 6 & {5} \\
        \hline
        DKINet & \makecell{\textbf{B01A},\textbf{A06A}, {N02B},\textbf{A12C},\\ \textbf{A01A},\textbf{A07A}, {N02A}, \textbf{C07A},\\ \textbf{B05C}, {N06A},\textbf{N05A}, \textbf{A02A}, \\ \textbf{A12A},  {N03A}, \textbf{A11D}} & \textbf{11} & 3 &  {4} \\
        \hline
        \hline
        2nd Diagnoses: & \multicolumn{4}{l}{\makecell{Z8673,F329,I471,F17210,Z9114,K760,B20,G92,R569,\\D6959,B181,E2740}}  \\
        \hline
        2nd Procedure:  & \multicolumn{4}{l}{\makecell{02HV33Z,009U3ZX} }\\
        \hline
        2nd Target: &\multicolumn{4}{l}{\makecell{G03A,A11D,J05A,B03B,N05A,A04A,N05B,A02A,A12A,\\C07A,A07A,A01A,B05C,A06A,A12C,B01A (16 codes)}} \\
        \hline
        Model & Predicted codes & \textbf{hit} & missed &  {unseen} \\
        \hline
        DKINet$_{w/o KG}$ & \makecell{\textbf{B01A},\textbf{A06A}, {N02B},\textbf{A12C},\\ \textbf{A01A},\textbf{A07A},\textbf{N05A},\textbf{A12A}, \\{C01E},\textbf{J05A}} & \textbf{8} & 8 &  {2} \\
        \hline
        DKINet & \makecell{\textbf{B01A},\textbf{A06A}, {N02B},\textbf{A12C},\\ \textbf{A01A},\textbf{A07A}, \textbf{N05B}, \textbf{C07A},\\ \textbf{B05C}, \textbf{N05A},\textbf{A12A},\textbf{A11D},\\ \textbf{J05A}} & \textbf{12} & 4 &  {1} \\
        \hline
    \end{tabular}
    
    \label{table6}
\end{table}

\section{Conclusion}
In this study, we proposed a new domain knowledge informed network to effectively incorporate the domain knowledge and clinical manifestation for medication recommendation, which is the first work incorporating the UMLS graph to recommend medications. Concretely, we first design a graph aggregation module to extract a more informative UMLS representation. Then, 
a knowledge-injected patient representation module is introduced to incorporate the domain knowledge based on the clinical manifestation of the patient. In addition, we designed a historical medication-aware patient representation module to capture the longitudinal dependency of historical medications on the current patient representation.
Extensive experimental results on the three publicly benchmark datasets have demonstrated the superiority of our proposed method.
\bibliography{arxiv_version}

\begin{thebibliography}{10}

\bibitem{almirall2012designing}
Daniel Almirall, Scott~N Compton, Meredith Gunlicks-Stoessel, Naihua Duan, and
  Susan~A Murphy.
\newblock Designing a pilot sequential multiple assignment randomized trial for
  developing an adaptive treatment strategy.
\newblock {\em Statistics in medicine}, 31(17):1887--1902, 2012.

\bibitem{UMLS}
Olivier Bodenreider.
\newblock The unified medical language system (umls): integrating biomedical
  terminology.
\newblock {\em Nucleic acids research}, 32(suppl\_1):D267--D270, 2004.

\bibitem{cao2019unifying}
Yixin Cao, Xiang Wang, Xiangnan He, Zikun Hu, and Tat-Seng Chua.
\newblock Unifying knowledge graph learning and recommendation: Towards a
  better understanding of user preferences.
\newblock In {\em The world wide web conference}, pages 151--161, 2019.

\bibitem{cheng2020club}
Pengyu Cheng, Weituo Hao, Shuyang Dai, Jiachang Liu, Zhe Gan, and Lawrence
  Carin.
\newblock Club: A contrastive log-ratio upper bound of mutual information.
\newblock In {\em International conference on machine learning}, pages
  1779--1788. PMLR, 2020.

\bibitem{retain}
Edward Choi, Mohammad~Taha Bahadori, Jimeng Sun, Joshua Kulas, Andy Schuetz,
  and Walter Stewart.
\newblock Retain: An interpretable predictive model for healthcare using
  reverse time attention mechanism.
\newblock {\em Advances in neural information processing systems}, 29, 2016.

\bibitem{dash2022review}
Tirtharaj Dash, Sharad Chitlangia, Aditya Ahuja, and Ashwin Srinivasan.
\newblock A review of some techniques for inclusion of domain-knowledge into
  deep neural networks.
\newblock {\em Scientific Reports}, 12(1):1040, 2022.

\bibitem{fitzgerald2009medication}
Richard~J FitzGerald.
\newblock Medication errors: the importance of an accurate drug history.
\newblock {\em British journal of clinical pharmacology}, 67(6):671--675, 2009.

\bibitem{gao2023medical}
Chao Gao, Shu Yin, Haiqiang Wang, Zhen Wang, Zhanwei Du, and Xuelong Li.
\newblock Medical-knowledge-based graph neural network for medication
  combination prediction.
\newblock {\em IEEE Transactions on Neural Networks and Learning Systems},
  2023.

\bibitem{hearst1998support}
Marti~A. Hearst, Susan~T Dumais, Edgar Osuna, John Platt, and Bernhard
  Scholkopf.
\newblock Support vector machines.
\newblock {\em IEEE Intelligent Systems and their applications}, 13(4):18--28,
  1998.

\bibitem{johnson2023mimic}
Alistair~EW Johnson, Lucas Bulgarelli, Lu~Shen, Alvin Gayles, Ayad Shammout,
  Steven Horng, Tom~J Pollard, Benjamin Moody, Brian Gow, Li-wei~H Lehman,
  et~al.
\newblock Mimic-iv, a freely accessible electronic health record dataset.
\newblock {\em Scientific data}, 10(1):1, 2023.

\bibitem{johnson2016mimic}
Alistair~EW Johnson, Tom~J Pollard, Lu~Shen, Li-wei~H Lehman, Mengling Feng,
  Mohammad Ghassemi, Benjamin Moody, Peter Szolovits, Leo Anthony~Celi, and
  Roger~G Mark.
\newblock Mimic-iii, a freely accessible critical care database.
\newblock {\em Scientific data}, 3(1):1--9, 2016.

\bibitem{kingma2014adam}
Diederik~P Kingma and Jimmy Ba.
\newblock Adam: A method for stochastic optimization.
\newblock {\em arXiv preprint arXiv:1412.6980}, 2014.

\bibitem{dmnc}
Hung Le, Truyen Tran, and Svetha Venkatesh.
\newblock Dual memory neural computer for asynchronous two-view sequential
  learning.
\newblock In {\em Proceedings of the 24th ACM SIGKDD International Conference
  on Knowledge Discovery \& Data Mining}, pages 1637--1645, 2018.

\bibitem{b5}
Deyin Liu, Yuanbo~Lin Wu, Xue Li, and Lin Qi.
\newblock Medi-care ai: Predicting medications from billing codes via robust
  recurrent neural networks.
\newblock {\em Neural Networks}, 124:109--116, 2020.

\bibitem{MCF}
Sicen Liu, Xiaolong Wang, Yang Xiang, Hui Xu, Hui Wang, and Buzhou Tang.
\newblock Multi-channel fusion lstm for medical event prediction using ehrs.
\newblock {\em Journal of Biomedical Informatics}, 127:104011, 2022.

\bibitem{b6}
Fenglong Ma, Yaqing Wang, Houping Xiao, Ye~Yuan, Radha Chitta, Jing Zhou, and
  Jing Gao.
\newblock A general framework for diagnosis prediction via incorporating
  medical code descriptions.
\newblock In {\em 2018 IEEE International Conference on Bioinformatics and
  Biomedicine (BIBM)}, pages 1070--1075. IEEE, 2018.

\bibitem{mcinnes2018umap}
Leland McInnes, John Healy, and James Melville.
\newblock Umap: Uniform manifold approximation and projection for dimension
  reduction.
\newblock {\em arXiv preprint arXiv:1802.03426}, 2018.

\bibitem{naumov2019deep}
Maxim Naumov, Dheevatsa Mudigere, Hao-Jun~Michael Shi, Jianyu Huang, Narayanan
  Sundaraman, Jongsoo Park, Xiaodong Wang, Udit Gupta, Carole-Jean Wu,
  Alisson~G Azzolini, et~al.
\newblock Deep learning recommendation model for personalization and
  recommendation systems.
\newblock {\em arXiv preprint arXiv:1906.00091}, 2019.

\bibitem{pollard2018eicu}
Tom~J Pollard, Alistair~EW Johnson, Jesse~D Raffa, Leo~A Celi, Roger~G Mark,
  and Omar Badawi.
\newblock The eicu collaborative research database, a freely available
  multi-center database for critical care research.
\newblock {\em Scientific data}, 5(1):1--13, 2018.

\bibitem{b3}
Alvin Rajkomar, Eyal Oren, Kai Chen, Andrew~M Dai, Nissan Hajaj, Michaela
  Hardt, Peter~J Liu, Xiaobing Liu, Jake Marcus, Mimi Sun, et~al.
\newblock Scalable and accurate deep learning with electronic health records.
\newblock {\em NPJ Digital Medicine}, 1(1):1--10, 2018.

\bibitem{read2011classifier}
Jesse Read, Bernhard Pfahringer, Geoff Holmes, and Eibe Frank.
\newblock Classifier chains for multi-label classification.
\newblock {\em Machine learning}, 85(3):333--359, 2011.

\bibitem{shang2019pre}
Junyuan Shang, Tengfei Ma, Cao Xiao, and Jimeng Sun.
\newblock Pre-training of graph augmented transformers for medication
  recommendation.
\newblock In {\em International Joint Conference on Artificial Intelligence}.
  International Joint Conferences on Artificial Intelligence, 2019.

\bibitem{gamenet}
Junyuan Shang, Cao Xiao, Tengfei Ma, Hongyan Li, and Jimeng Sun.
\newblock Gamenet: Graph augmented memory networks for recommending medication
  combination.
\newblock In {\em proceedings of the AAAI Conference on Artificial
  Intelligence}, volume~33, pages 1126--1133, 2019.

\bibitem{sun2022debiased}
Hongda Sun, Shufang Xie, Shuqi Li, Yuhan Chen, Ji-Rong Wen, and Rui Yan.
\newblock Debiased, longitudinal and coordinated drug recommendation through
  multi-visit clinic records.
\newblock {\em Advances in Neural Information Processing Systems},
  35:27837--27849, 2022.

\bibitem{wang2019multi}
Hongwei Wang, Fuzheng Zhang, Miao Zhao, Wenjie Li, Xing Xie, and Minyi Guo.
\newblock Multi-task feature learning for knowledge graph enhanced
  recommendation.
\newblock In {\em The world wide web conference}, pages 2000--2010, 2019.

\bibitem{wang2021learning}
Xiang Wang, Tinglin Huang, Dingxian Wang, Yancheng Yuan, Zhenguang Liu,
  Xiangnan He, and Tat-Seng Chua.
\newblock Learning intents behind interactions with knowledge graph for
  recommendation.
\newblock In {\em Proceedings of the Web Conference 2021}, pages 878--887,
  2021.

\bibitem{ARMR}
Yanda Wang, Weitong Chen, Dechang Pi, and Lin Yue.
\newblock Adversarially regularized medication recommendation model with
  multi-hop memory network.
\newblock {\em Knowledge and Information Systems}, 63(1):125--142, 2021.

\bibitem{wu2022leveraging}
Jialun Wu, Buyue Qian, Yang Li, Zeyu Gao, Meizhi Ju, Yifan Yang, Yefeng Zheng,
  Tieliang Gong, Chen Li, and Xianli Zhang.
\newblock Leveraging multiple types of domain knowledge for safe and effective
  drug recommendation.
\newblock In {\em Proceedings of the 31st ACM International Conference on
  Information \& Knowledge Management}, pages 2169--2178, 2022.

\bibitem{COGNet}
Rui Wu, Zhaopeng Qiu, Jiacheng Jiang, Guilin Qi, and Xian Wu.
\newblock Conditional generation net for medication recommendation.
\newblock In {\em Proceedings of the ACM Web Conference 2022}, pages 935--945,
  2022.

\bibitem{yangmanydg}
Chaoqi Yang, M~Brandon Westover, and Jimeng Sun.
\newblock Manydg: Many-domain generalization for healthcare applications.
\newblock In {\em The Eleventh International Conference on Learning
  Representations}, 2022.

\bibitem{MICRON}
Chaoqi Yang, Cao Xiao, Lucas Glass, and Jimeng Sun.
\newblock Change matters: Medication change prediction with recurrent residual
  networks.
\newblock In {\em 30th International Joint Conference on Artificial
  Intelligence, IJCAI 2021}, pages 3728--3734. International Joint Conferences
  on Artificial Intelligence, 2021.

\bibitem{safedrug}
Chaoqi Yang, Cao Xiao, Fenglong Ma, Lucas Glass, and Jimeng Sun.
\newblock Safedrug: Dual molecular graph encoders for recommending effective
  and safe drug combinations.
\newblock In {\em Proceedings of the Thirtieth International Joint Conference
  on Artificial Intelligence}, 2021.

\bibitem{yang2021gfe}
Zuoxi Yang, Shoubin Dong, and Jinlong Hu.
\newblock Gfe: General knowledge enhanced framework for explainable sequential
  recommendation.
\newblock {\em Knowledge-Based Systems}, 230:107375, 2021.

\bibitem{yin2019domain}
Changchang Yin, Rongjian Zhao, Buyue Qian, Xin Lv, and Ping Zhang.
\newblock Domain knowledge guided deep learning with electronic health records.
\newblock In {\em 2019 IEEE International Conference on Data Mining (ICDM)},
  pages 738--747. IEEE, 2019.

\bibitem{yu2021ernie}
Fei Yu, Jiji Tang, Weichong Yin, Yu~Sun, Hao Tian, Hua Wu, and Haifeng Wang.
\newblock Ernie-vil: Knowledge enhanced vision-language representations through
  scene graphs.
\newblock In {\em Proceedings of the AAAI Conference on Artificial
  Intelligence}, volume~35, pages 3208--3216, 2021.

\bibitem{leap}
Yutao Zhang, Robert Chen, Jie Tang, Walter~F Stewart, and Jimeng Sun.
\newblock Leap: learning to prescribe effective and safe treatment combinations
  for multimorbidity.
\newblock In {\em proceedings of the 23rd ACM SIGKDD international conference
  on knowledge Discovery and data Mining}, pages 1315--1324, 2017.

\end{thebibliography}
\end{document}